\documentclass{article}

\PassOptionsToPackage{numbers}{natbib}
\usepackage[preprint]{neurips_2025}

\usepackage[utf8]{inputenc}
\usepackage[T1]{fontenc}
\usepackage{hyperref}
\usepackage{url}
\usepackage{booktabs}
\usepackage{amsfonts}
\usepackage{nicefrac}
\usepackage{microtype}
\usepackage{xcolor} 
\usepackage{caption}
\usepackage{subcaption}
\usepackage[pdftex]{graphicx}
\usepackage{tikz}
\usepackage{times}
\usepackage{tabularx}
\usepackage{float}
\usepackage{comment}
\usepackage{listings}
\usepackage{epstopdf}
\usepackage{fancyhdr}
\usepackage{lipsum}
\usepackage{multirow}
\usepackage{amsmath}
\usepackage{amssymb}
\usepackage{mathtools}
\usepackage{amsthm}

\theoremstyle{plain}
\newtheorem{theorem}{Theorem}[section]

\theoremstyle{definition}

\theoremstyle{remark}

\newcommand{\R}{\mathbb{R}}

\newcommand{\E}{\mathbb{E}}

\title{Discretized Quadratic Integrate-and-Fire Neuron Model for Deep Spiking Neural Networks}

\author{
  Eric~Jahns \\
  STAM Center\\
  Arizona State University\\
  Tempe, Arizona \\
  \texttt{jjahns@asu.edu} \\
  \And
  Davi Moreno \\
  Center for Advanced Studies and \\
  Systems of Recife\\
  \texttt{dcma@cesar.org.br} \\
  \And
  Milan Stojkov \\
  Faculty of Technical Sciences \\ 
  University of Novi Sad \\
  Novi Sad, Serbia \\
  \texttt{stojkovm@uns.ac.rs} \\
  \And
  Michel A. Kinsy \\
  STAM Center\\
  Arizona State University\\
  Tempe, Arizona \\
  \texttt{mkinsy@asu.edu} \\
}

\begin{document}

\maketitle

\begin{abstract}

Spiking Neural Networks (SNNs) have emerged as energy-efficient alternatives to traditional artificial neural networks, leveraging asynchronous and biologically inspired neuron dynamics. Among existing neuron models, the Leaky Integrate-and-Fire (LIF) neuron has become widely adopted in deep SNNs due to its simplicity and computational efficiency. However, this efficiency comes at the expense of expressiveness, as LIF dynamics are constrained to linear decay at each timestep. In contrast, more complex models, such as the Quadratic Integrate-and-Fire (QIF) neuron, exhibit richer, nonlinear dynamics but have seen limited adoption due to their training instability. On that note, we propose the first discretization of the QIF neuron model tailored for high-performance deep spiking neural networks and provide an in-depth analysis of its dynamics. To ensure training stability, we derive an analytical formulation for surrogate gradient windows directly from our discretizations' parameter set, minimizing gradient mismatch. We evaluate our method on CIFAR-10, CIFAR-100, ImageNet, and CIFAR-10 DVS, demonstrating its ability to outperform state-of-the-art LIF-based methods. These results establish our discretization of the QIF neuron as a compelling alternative to LIF neurons for deep SNNs, combining richer dynamics with practical scalability.

\end{abstract}

\section{Introduction}

Artificial Neural Networks (ANNs) have gained mainstream adoption in recent years, thanks to their success in domains such as computer vision \citep{JunyiChai-2021} and natural language processing \citep{WahabKhan-2023}. However, the energy demands of ANNs continue to grow \citep{KashuYamazaki-2022}. In contrast, Spiking Neural Networks (SNNs) have gained attention as an energy-efficient alternative. Unlike traditional ANNs, which process continuous-valued data synchronously, SNNs operate asynchronously on discrete events known as spikes. These spikes are driven by biologically inspired neuron dynamics, allowing SNNs to replicate the brain's sparse connectivity and energy-efficient structure \citep{WulframGerstner-2014}. As a result, when SNNs are implemented on hardware tailored to these characteristics, they can operate with lower energy consumption than traditional ANN models \citep{NitinRathi-2023}. This type of hardware is typically referred to as neuromorphic hardware. Examples of SNNs implemented on neuromorphic hardware can be seen in applications such as utilizing IBM TrueNorth \citep{FilippAkopyan-2015} for always-on speech recognition for edge devices \citep{Wei-YuTasi-2017} or leveraging Intel Loihi 2 \citep{IntelLoihi2} for ultra-low-power image classification \citep{GregorLenz-2023}.

In deep learning with SNNs, the Leaky Integrate-and-Fire (LIF) neuron model has become the standard due to its unique blend of biologically realistic dynamics and computational efficiency \citep{WulframGerstner-2014, YujieWu-2018}. However, this computational efficiency comes at a cost, as the LIF neuron model's dynamics are limited to linear decay proportional to its voltage. While other neuron models, such as the Quadratic Integrate-and-Fire, Exponential Integrate-and-Fire, Izhikevich, and Hodgkin-Huxley, exhibit more complex dynamics~\citep{WulframGerstner-2014}, they incur increased computational costs and training instability when utilized in deep SNNs. Interestingly, the Quadratic Integrate-and-Fire (QIF) neuron strikes a balance with more expressive dynamics than the LIF neuron and minimal computational overhead compared to other complex neuron models. Prior attempts to integrate QIF neurons \citep{YunJhuLee-2021, WenChiehWu-2021, WeijieYe-2021} have been restricted to shallow networks or niche applications. Consequently, incorporating the QIF neuron into scalable deep SNNs remains an open and underexplored challenge.

To address this challenge, we propose a novel discretization of the QIF neuron, which exhibits biologically inspired dynamics and proves to be well-suited for deep SNNs. Our discretization of the QIF neuron enables the representation of more complex dynamics, such as subthreshold oscillations and sensitivity to input variations, which are not seen in LIF neurons' simple integration and leakage. To the best of our knowledge, this is the first demonstration of deep SNNs based on QIF neurons. The main contributions of this work are summarized as follows:

\begin{itemize}
    \item We introduce a novel discretization of the QIF neuron model that captures key nonlinear dynamics, such as subthreshold oscillations and input variation sensitivity, absent in traditional LIF models. Our formulation preserves biological plausibility while enabling efficient integration into deep SNN architectures.
    \item To ensure stable and effective training, we derive an analytical surrogate gradient window directly from the discretized QIF's parameterization. This principled approach minimizes gradient mismatch and eliminates the need for heuristically tuned surrogate functions.
    \item Our method achieves state-of-the-art accuracy among LIF-based convolutional deep SNNs on both static and neuromorphic datasets, with results of $96.86\%$ on CIFAR-10, $80.62\%$ on CIFAR-100, and $80.70$\% on CIFAR10-DVS using ResNet-19, surpassing the previous best results by $0.04\%$, $0.42\%$, and $0.86$\%, respectively. We also achieve competitive performance on ImageNet with $70.52\%$ accuracy, demonstrating our method's scalability. Despite its enhanced dynamics, our QIF neuron discretization incurs only minimal energy overheads, ranging from $0.94\%$ to $3.23\%$ compared to LIF neurons.
\end{itemize}

\section{Related Work}

\subsection{Deep Learning with Spiking Neural Networks}

Training SNNs directly with backpropagation is challenging due to the non-differentiability of spikes \citep{ZexiangYi-2023}. Several techniques have been developed to overcome this, with one of the most widely adopted being surrogate gradients. Surrogate gradients attempt to approximate the derivative of a spike with respect to the membrane potential using a differentiable function \citep{YujieWu-2018}. In addition to surrogate gradients, works employ various techniques to improve direct training performance. Some of these techniques include batch or membrane potential normalization \citep{HanleZheng-2020, ChaotengDuan-2022, YufeiGuo-2023}, developing new loss functions \citep{ShikuangDeng-2022, YufeiGuo-2022}, learning surrogate gradient behavior \citep{YuhangLi-2021, ShuangLian-2023, ShikuangDeng-2023}, and more. Due to the lack of support for training on neuromorphic datasets when using ANN-to-SNN conversion techniques~\citep{YongqiangCao-2015}, we restrict any comparisons to direct training techniques.

\subsection{Neuron Models and Parameter Learning}

Several works that utilize direct training techniques, such as \citet{WeiFang-2021, XingtingYao-2022, NitinRathi-2023, ShuangLian-2023, ShuangLian-2024}, modify the LIF neuron by either altering its dynamics or incorporating learnable parameters. \citet{WeiFang-2021} proposes a learnable decay factor for the LIF neuron, which can be independently optimized for each layer. \citet{NitinRathi-2023} takes this a step further by co-optimizing the decay and threshold of each spiking layer. \citet{XingtingYao-2022} proposed gating features, similar to long-short term memory, that can choose between various biological features implemented in their model. \citet{ShuangLian-2023} propose using a learnable decay parameter to dynamically adjust the surrogate gradient window, thereby fitting the LIF neuron's voltage distribution throughout the training process. \citet{ShuangLian-2024} proposes using a temporal-wise attention mechanism to selectively establish connections between current and past temporal data. \citet{YufeiGuo-2024} introduced the notion of ternary spikes to reduce information loss. While the LIF neuron has improved and showcased promising performance in many deep-learning applications, its dynamics are fundamentally constrained to linear decay proportional to the neuron's voltage. Inspired by this, we look towards other neuron models to quantify and address this limitation.

\section{Background}

\subsection{Spiking Neural Networks}

While ANNs use continuous-valued data to transmit information, SNNs use discrete events called spikes. In modern deep SNN research, the LIF neuron is the most widely adopted for its computational efficiency and biologically plausible dynamics \citep{WulframGerstner-2014}. The LIF neurons' dynamics are defined by
\begin{align}
    \label{eq:LIFODE}
    \tau \frac{du}{dt} = u_{r} - u + RI,
\end{align}
where $\tau$ is a membrane time constant, $u$ is the membrane potential, $u_{r}$ is the resting potential, $R$ is a linear resistor, and $I$ is pre-synaptic input. When using the LIF neuron in deep learning scenarios, discretization is required \citet{ChaotengDuan-2022}. The most commonly used discretization was introduced by \citet{YujieWu-2018}, who utilized Euler's method to solve~\eqref{eq:LIFODE}. They defined their model as
\begin{align}
    \label{eq:LIF}
    u(t+1) = \beta u(t) + I(t).
\end{align}
In~\eqref{eq:LIF}, $t$ denotes the current timestep, $\beta$ is a membrane potential decay factor, $u$ is the membrane potential of a neuron, and $I$ are pre-synaptic inputs into the neuron. Given a threshold, $u_{th}$, when $u(t) > u_{th}$, a spike is produced and is denoted $o(t+1)$. \citet{YujieWu-2018} further defines an iterative update rule for both spatial and temporal domains as
\begin{align}
    \label{eq:LIFUpdate1}
    u(t+1) &= \beta u(t)(1 - o(t)) + I(t) \\
    \label{eq:LIFUpdate2}
    o(t+1) &= \Theta(u(t+1) - u_{th}),
\end{align}
where $\Theta$ is the Heaviside function given by $\Theta(x) = 0$ if $x < 0$, else $\Theta(x) = 1$.

Both the ordinary differential equation, shown in~\eqref{eq:LIFODE}, and the discretized equation, shown in~\eqref{eq:LIF}, of the LIF neuron model are constrained to a linear decay directly proportional to the voltage.

\subsection{Quadratic Integrate-and-Fire Neuron Model}

The Hodgkin-Huxley (HH) neuron model was created to mimic the activity of neurons found in a giant squid and has proven itself invaluable in the field of neuroscience \citep{WulframGerstner-2014}. Over the years, simplifications of the HH neuron have been introduced to reduce its large computational complexity. The LIF neuron is an extreme example of this. However, unlike the HH neuron's non-linear voltage dynamics, the LIF neuron utilizes linear decay proportional to voltage. The Quadratic Integrate-and-Fire (QIF) neuron model is another simplification that maintains the non-linear dynamics of the HH neuron. However, this comes at the cost of increased computational complexity compared to the LIF neuron. As described in~\citep{WulframGerstner-2014}, the dynamics of the QIF neuron are defined as
\begin{equation}
    \label{eq:QIFODE}
    \tau \frac{du}{dt} = a(u - u_{r})(u - u_c) + RI,
\end{equation}
where $\tau$ is a membrane time constant, $a$ is a sharpness parameter controlling the rate of decay, $u$ is the membrane potential, $u_{r}$ is the resting potential, $u_c$ is the critical spiking threshold, $R$ is a resistance, and $I$ is the pre-synaptic input. Additionally, it must hold that $a > 0$ and $u_{r} < u_c$. Unlike the LIF neuron model, the QIF neuron model contains non-linear voltage dynamics proportional to the voltage's square. This enables the QIF neuron to exhibit varying dynamics depending on its current voltage. For example, the QIF neuron can decay rapidly when $u < u_{th}$, or increase rapidly as $u$ approaches and exceeds $u_c$ \citep{WulframGerstner-2014}.

\subsection{Threshold-Dependent Batch Normalization}
\citet{SergeyIoffe-2015} first introduced the concept of batch normalization to accelerate ANN training by reducing the internal covariant shift of each layer. Batch normalization was originally designed to normalize spatial data, not spatial-temporal data. On this note, \citet{HanleZheng-2020} proposed threshold-dependent Batch Normalization (tdBN), which normalizes pre-synaptic input, $I$, in both the spatial and temporal domains based on the neuron's threshold, $u_{th}$. Suppose $I_k(t)$ represents the $k_{th}$ feature map of $I$ at timestep $t$. Then, we normalize each feature map $I_k = (I_k(1), I_k(2), \ldots, I_k(T))$ in the temporal domain by
\begin{align}
    \label{eq:tdBN}
    \hat{I}_k &= \frac{\eta u_{th}(I_k - \E[I_k])}{\sqrt{\mathrm{Var}(I_k) + \epsilon}}, \quad \Bar{I}_k = \gamma \hat{I_k} + \xi,    
\end{align}
where $\E$ and $\mathrm{Var}$ compute the mean and variance of $I_k$, $\eta$ is used for residual connections, $\epsilon$ is a small positive real number, and $\gamma$ and $\xi$ are learnable parameters. Following tdBN, $\hat{I} \sim \mathcal{N}(0, u_{th}^2)$.

\subsection{Training Spiking Neural Networks}

The Spatial-Temporal Back Propagation (STBP) algorithm is commonly used to train SNNs~\citep{YujieWu-2018}. First, an SNN processes temporal data for $T$ timesteps. The SNN's output is decoded by accumulating the synaptic voltage of the last layer as follows
\begin{align}
    \hat{y} = \frac{1}{T} \sum_{t=1}^T W o(t).
\end{align}
In the above equation, $W$ is a weight matrix, $o(t)$ is the spiking activity at timestep $t$, and $\hat{y} \in \mathbb{R}^m$ is the model output given $m$ classes. Using our output vector $\hat{y} = (\hat{y}_1, \hat{y}_2, \ldots, \hat{y}_m)$ and a label vector $y = (y_1, y_2, \ldots, y_m)$, we compute the cross entropy loss, $L$, between $\hat{y}$ and $y$. Then, using the STBP algorithm, we train the network using the chain rule to update synaptic weights by
\begin{align}
    \frac{\partial L}{\partial W_n} &= \sum_{t=1}^T 
    \left(
        \frac{\partial L}{\partial o_n(t)} \frac{\partial o_n(t)}{\partial u_n(t)} 
        + \frac{\partial L}{\partial u_n(t+1)}\frac{\partial u_n(t+1)}{\partial u_n(t)}
    \right) \times \frac{\partial u_n(t)}{\partial W_n},
\end{align}
where $n$ is the layer of the network, $u(t)$ is a neuron's membrane potential at timestep $t$, and $o(t)$ is a neuron's spike output at timestep $t$ \citep{YufeiGuo-2023}.

To overcome the undefined derivative, $\frac{\partial o_n(t)}{\partial u_n(t)}$, \citet{YujieWu-2018} proposed using the derivative of an approximation to the Heaviside function with useful gradient information. This technique is known as a \textbf{surrogate gradient}. One common surrogate gradient is the rectangle function, as seen in \citet{HanleZheng-2020, ShikuangDeng-2022, ShuangLian-2023}, and is defined by
\begin{align}
    \label{eq:RectangleSG}
    \frac{\partial o(t)}{\partial u(t)} \approx \frac{1}{\alpha} \mathrm{sign}\left (|u(t) - u_{th}| < \frac{\alpha}{2}\right ),
\end{align}
where $\alpha$ determines the width and area of the surrogate gradient and typically remains constant throughout training. The choice of $\alpha$ has a significant impact on the learning process of SNNs, with improper choices leading to gradient mismatches and approximation errors.

\section{Methodology}

\subsection{Discretized Quadratic Integrate-and-Fire Neuron Model}

To effectively utilize the QIF neuron model within a deep learning setting, discretization is required \citep{ChaotengDuan-2022}. By applying Euler's method to discretize the QIF neuron model in~\eqref{eq:QIFODE} and performing some simplifications (see Appendix~\ref{appendix:discretization}), we obtain

\begin{align}
    \label{eq:eq:DiscretizedQIFSimplified}
    u(t+1) = a(u(t)-u_{r})(u(t)-u_c) + u(t) + I(t).
\end{align}
To reduce computational complexity—by eliminating one addition—and to facilitate the dynamic analysis of the model,~\eqref{eq:eq:DiscretizedQIFSimplified} is rewritten as
\begin{align}
    \label{eq:QIF}
    u(t+1) = a(u(t)-u_1)(u(t)-u_2) + I(t),
\end{align}
where $u_1$ and $u_2$ can be computed from $a$, $u_r$, and $u_c$ (see Appendix~\ref{appendix:discretization}), by using
\begin{equation}
\begin{aligned}
    \label{eq:QIF_param_calc}
    u_{1,2} &= \frac{1}{2}\left[(u_r+u_c) - \frac{1}{a}\right] \mp \frac{1}{2} \sqrt{\Delta}\\
    \Delta &= (u_c-u_r)^2 - \frac{2(u_r+u_c)}{a} + \frac{1}{a^2}.
\end{aligned}
\end{equation}

When incorporating our discretization of the QIF neuron~\eqref{eq:QIF} into existing deep SNN architectures, we adopt and modify the iterative update rule proposed by~\citet{YujieWu-2018} to obtain
\begin{align}
    \label{eq:QIFUpdate}
    f_{n}(t) &= a(u_{n}(t)-u_{1})(u_{n}(t)-u_{2})  \\
    u_{n}(t+1) &= f_{n}(t)(1 - o_{n}(t)) + u_\text{reset}o_{n}(t) + I_{n}(t) \\
    o_{n}(t+1) &= \Theta(u_{n}(t+1) - u_{th}).
\end{align}
In the above equations, $t$ denotes the timestep, $n$ denotes the layer of the network, $I$ is pre-synaptic input, $u$ is the membrane potential, $u_{th}$ is the firing threshold, and $\Theta$ is the Heaviside function. When the membrane potential exceeds the firing threshold $u_{th}$, a spike will be produced and its potential will be reset to $u_\text{reset}$. As in the LIF case~\eqref{eq:LIFUpdate1}, in which a spike resets the membrane potential to zero, we also use $u_\text{reset}=0$ in this work.

\subsection{Model Dynamics and Parameter Selection}

\begin{figure}
\centering
\begin{subfigure}{.5\textwidth}
  \centering
  \includegraphics[width=\linewidth]{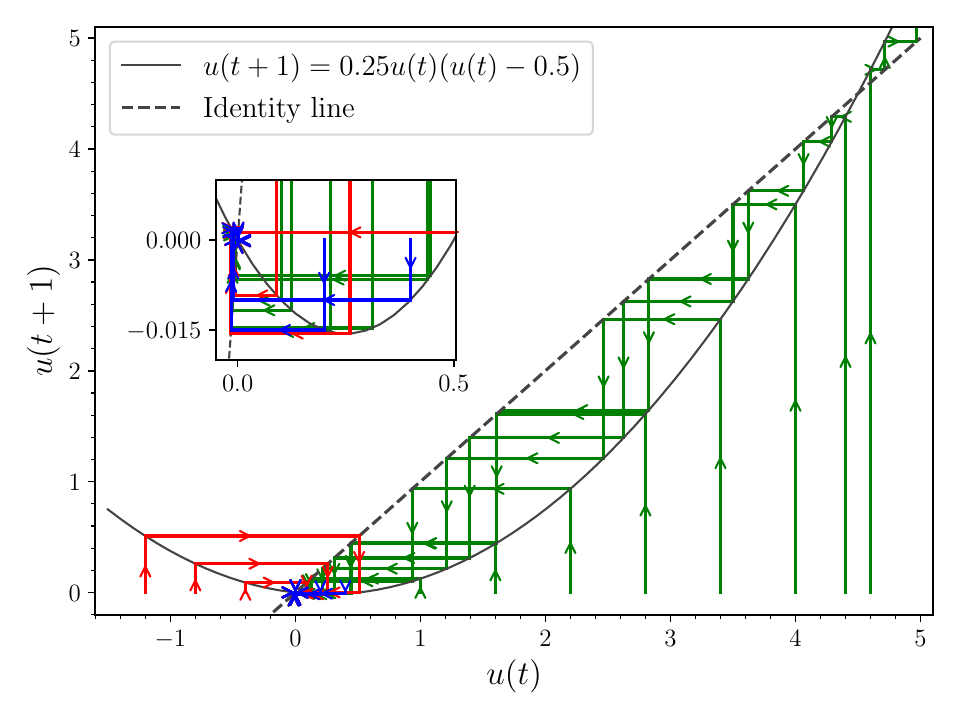}
  \caption{}
  \label{fig:cobweb}
\end{subfigure}%
\begin{subfigure}{.5\textwidth}
  \centering
  \includegraphics[width=\linewidth]{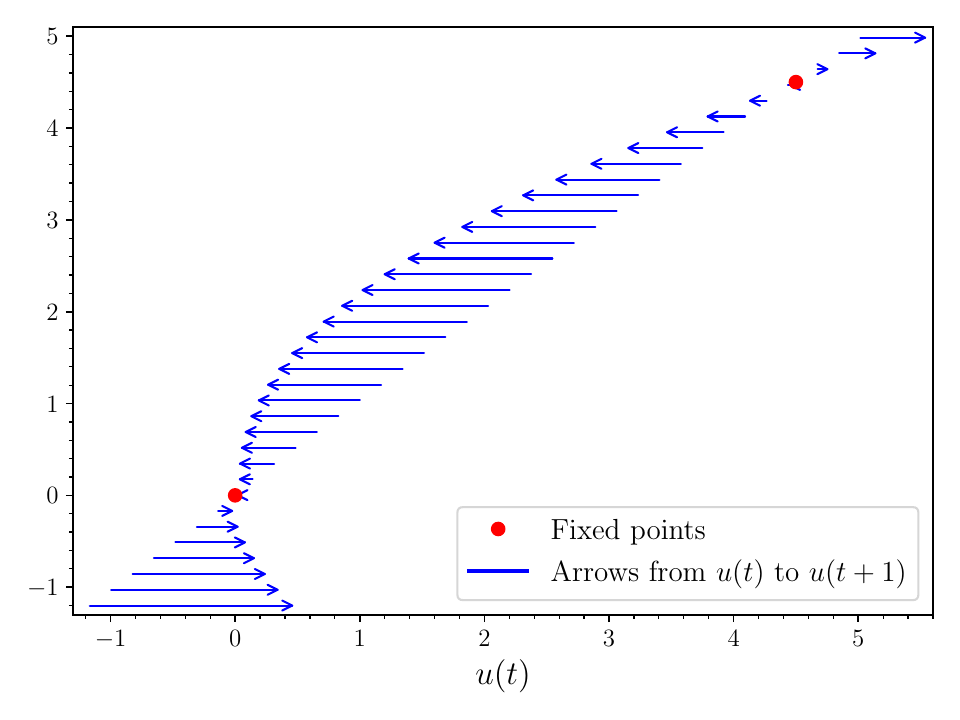}
  \caption{}
  \label{fig:phase_portrait}
\end{subfigure}
\caption{Dynamics of our discretized QIF in~\eqref{eq:QIF} with $a=0.25$, $u_1=0$, and $u_2=0.5$. (a) Cobweb plot and (b) phase portrait for different initial conditions. The $y$-axis in (b) is used solely to aid in visualizing the arrows.}
\label{fig:test}
\end{figure}

To perform a dynamical analysis of our QIF discretization, we first consider the recurrence relation in~\eqref{eq:eq:DiscretizedQIFSimplified} with zero input, i.e., $I(t) = 0$. By setting $u(t+1) = u(t)$, 
\begin{align}
    0 = a(u(t)-u_{r})(u(t)-u_c),
\end{align}
it follows that both $u_r$ and $u_c$ are fixed points of the equation. The stability of the fixed points is determined by evaluating the derivative $\frac{du(t+1)}{du(t)}$ at those points. By differentiating~\eqref{eq:QIF}, we obtain: 
\begin{equation}
\label{eq:QIF_derivative}
    g(u(t)) = \frac{du(t+1)}{du(t)} = 2au(t) - a(u_1 + u_2).
\end{equation} 
A fixed point $u^*$ is considered stable if the condition $|g(u^*)| < 1$ is satisfied, and unstable if $|g(u^*)| > 1$. A stable fixed point at zero is of particular interest, as it attracts a range of initial conditions near the origin, a behavior also observed in the LIF neuron, where the voltage gradually decays to zero over time due to leakage. With this in mind, $u_r = 0$ was chosen, simplifying~\eqref{eq:QIF_param_calc} to
\begin{equation}
\begin{aligned}
    \label{eq:QIF_param_calc_ur=0}
    u_1 = 0, \quad
    u_2 = u_c - \frac{1}{a} \longrightarrow u_2 > -\frac{1}{a} ,\\
\end{aligned}
\end{equation}
where the restriction for $u_2$ comes from the constraint $u_c>u_r$ in~\eqref{eq:QIFODE}. The parameter $a$ can be interpreted as a gain or leakage factor, similar to $\beta$ in the LIF neuron~\eqref{eq:LIFUpdate1}. Based on $\beta$ values commonly used in other LIF-related studies~\citep{YufeiGuo-2024, ShuangLian-2023, WeiFang-2021}, a value of $a = \frac{1}{4} = 0.25$ is selected, restricting $u_2>-4$ in~\eqref{eq:QIF_param_calc_ur=0}.
Given the values $a=0.25$ and $u_1=0$, we chose the value $u_2=0.5$, and use it to analyze the dynamics of~\eqref{eq:QIF}. For this choice of parameters, it follows from~\eqref{eq:QIF_param_calc_ur=0} that
\begin{equation} 
    u_c = \frac{1}{a} + u_2 = 4 + 0.5 = 4.5. 
\end{equation} 
Therefore, the evaluation of~\eqref{eq:QIF_derivative} at the fixed points is given by
\begin{align} 
    g(u_r)&=g(0)=-\frac{1}{8} \longrightarrow |g(0)| < 1 \text{ (stable)} \\
    g(u_c)&=g(4.5)=\frac{17}{8} \longrightarrow |g(4.5)| > 1 \text{ (unstable)}, 
\end{align} 
and so, zero is identified as a stable fixed point, as desired, while $4.5$ is classified as an unstable fixed point. For this choice of parameters and under zero input, an example plot of our discretization in~\eqref{eq:QIF} is shown in Figure~\ref{fig:qif_plot}. The colored regions are used to indicate the zones in which the model exhibits distinct types of behavior, described as follows:

\begin{figure}
  \centering
  \includegraphics[width=0.65\linewidth]{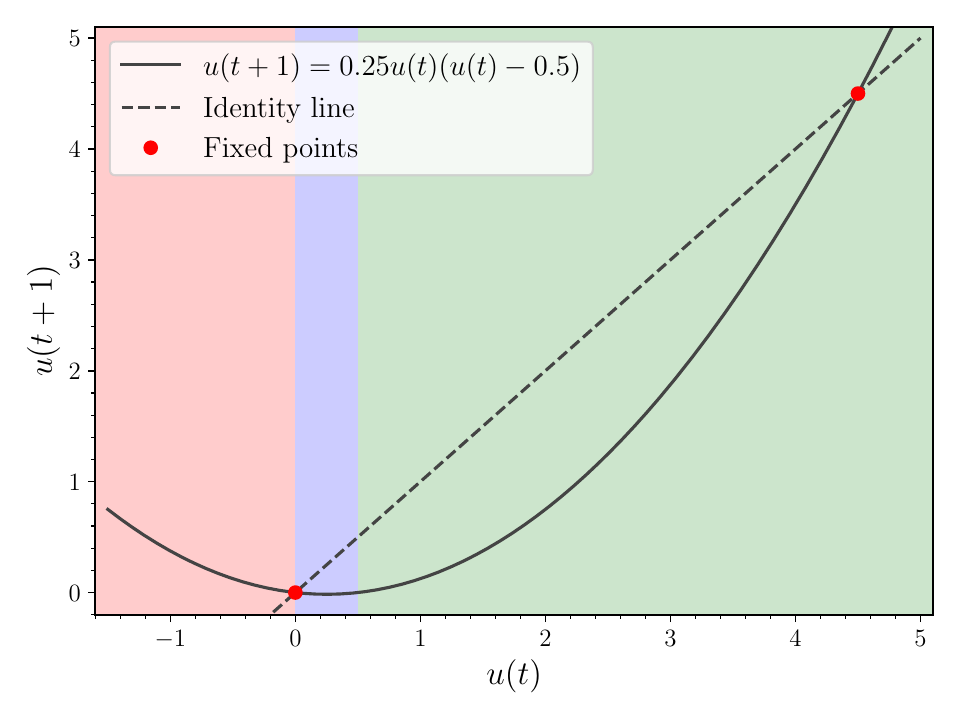}
  \caption{Plot of our discrete QIF recurrence relation~\eqref{eq:QIF} for null input and parameters $a = 0.25$, $u_1 = 0$, and $u_2 = 0.5$.}
  \label{fig:qif_plot}
\end{figure}
\begin{itemize}
    \item 
    \textbf{Green region ($u(t) > \max(u_1, u_2)$):} In this region, if $u(t) \leq u_c$, then $u(t+1) \leq u(t)$, indicating that a nonlinear factor attenuates the membrane potential. Conversely, if $u(t) > u_c$, the opposite occurs, and a nonlinear gain causes $u(t+1) > u(t)$. This latter behavior is undesirable and caused by the instability of the fixed point $u_c$. To prevent this in our discretization, the voltage threshold $u_{th}$ is constrained as follows: 
    \begin{align} 
        \label{eq:uth_constrain}
        \max(u_1, u_2) \leq u_{th} \leq u_c,
    \end{align} 
    such that whenever the membrane potential crosses this threshold, it is reset and a spike is generated. This ensures that only the attenuating (amortizing) behavior is exhibited. This behavior resembles the leakage observed in the LIF model, in which the membrane potential integrates past values, subject to a leakage factor.

    \item \textbf{Blue region ($\min(u_1, u_2) \leq u(t) \leq \max(u_1, u_2)$):} In this region, $u_\text{min} \leq u(t+1) \leq 0$, where $u_\text{min} = -\frac{a}{4}(u_1 - u_2)^2$, as shown in Appendix~\ref{appendix:discretization}. For the chosen parameters, this minimum value is small ($u_\text{min} = -\frac{1}{64}$). In this region, the neuron exhibits subthreshold oscillations, which alone are insufficient to trigger action potentials without additional input current. Nonetheless, subthreshold oscillations found in biological neurons are thought to aid in the processing of sensory inputs~\citep{IlanLampl-1993}.

    \item \textbf{Red region ($u(t) < \min(u_1, u_2)$):} This region also activates neurons that would typically be considered inactive in LIF models, by transforming a negative voltage $u(t)$ into a positive $u(t+1)$. As a result, a negative $u(t)$ followed by a sufficiently positive input $I(t)$ can lead to a spike. This phenomenon replicates one observed in which specific biological neurons can spike in response to significant changes in their input. For instance, this kind of input-sensitive spiking behavior has been reported in specific sensory neurons, such as retinal ganglion cells, which can fire in response to rapid changes in light intensity~\citep{Nirenberg1997}. This behavior is not captured by the LIF model~\eqref{eq:LIF}, which only includes integration and leakage components.
\end{itemize}
To illustrate the dynamic evolution of our QIF discretization~\eqref{eq:QIF} under different initial conditions within each of the described colored regions, a Cobweb plot is presented in Figure~\ref{fig:cobweb}. It can be observed that the fixed point at zero attracts nearby initial conditions, while the fixed point at 4.5 causes nearby initial conditions to diverge from it, as expected. A phase portrait is also provided in Figure~\ref{fig:phase_portrait}, depicting the same behavior at the fixed points. It should be noted that negative initial conditions are mapped to positive values after one iteration, illustrating the behavior described for the red region in Figure~\ref{fig:qif_plot}.

This work demonstrates that learning can be achieved using models that mimic the behavior of biological neurons, extending beyond simple integration with leakage, as seen in the LIF model~\eqref{eq:LIF}. By analyzing the colored regions in Figure~\ref{fig:qif_plot}, it was shown that our discretization of the QIF model can represent not only integration with leakage, but also subthreshold oscillations and sensitivity to variations in the neuron's input. Since a threshold value of $u_{th} = 0.5$ is adopted in related works ~\citep{YufeiGuo-2024, ShuangLian-2023, WeiFang-2021}, the same value was chosen for consistency. From the chosen parameters, we have that $u_{th}=u_2=0.5$. This emphasizes the relevance of the red and blue regions from Figure~\ref{fig:qif_plot} in learning, as shown by the results in Section~\ref{sec:experiments}. To summarize, the final set of parameters used in our discretization of the QIF neuron model~\eqref{eq:QIF} is $a = 0.25$, $u_1 = 0$, $u_2 = 0.5$, and $u_{th} = 0.5$.

\subsection{Surrogate Gradient Formulation}

Our discretized QIF neuron~\eqref{eq:QIF} produces voltage distributions whose mean and variance widely change based on the chosen parameter set. Therefore, na\"ivley applying the surrogate gradient in \eqref{eq:RectangleSG} can lead to severe gradient mismatch. To remedy this issue, we propose an analytical means to calculate the surrogate gradient window based on the parameter set of our QIF discretization. More specifically, we assume that all pre-synaptic input is normalized with tdBN~\eqref{eq:tdBN} such that $I \sim \mathcal{N}(0, u_{th}^2)$ and propose Theorem \ref{thm:1} to explain the statistical properties of our QIF neuron model.
\begin{theorem}
    \label{thm:1}
    Under our discrete QIF neuron model using tdBN to normalize pre-synaptic input $I$, such that $I \sim \mathcal{N}(0, u_{th}^2)$, the membrane potential $u$ follows a distribution with mean $\mu_u~\approx a f(u_{th}, u_1, u_2)$ and variance $\sigma_u^2 \approx u_{th}^2 h(u_{th}, u_1, u_2, a)$ where $\mu_u$ and $\sigma_u^2$ are directly proportional to the functions $f$ and $h$, respectively. The functions $f$ and $h$ are defined as $f(u_{th}, u_1, u_2) = u_{th}^2 + u_1u_2$ and $h(u_{th}, u_1, u_2, a) = 1 + a^2 (2u_{th}^2 + (u_1 + u_2)^2)$.
\end{theorem}
The proof of Theorem \ref{thm:1} can be seen in Appendix \ref{appendix:proofs}. Theorem \ref{thm:1} states that after integrating tdBN normalized pre-synaptic inputs into our discrete QIF neuron according to~\eqref{eq:QIF}, the membrane potential $u$ follows a distribution with mean $\mu_u$ and variance $\sigma_u^2$. Therefore, we can approximate the values of $\mu_{u}$ and $\sigma_{u}^2$ using Theorem \ref{thm:1} to calculate the surrogate gradient window based on the parameters $a, u_{th}, u_1,$ and $u_2$, reducing the risk of poor window choice and potential gradient mismatch. We define our new surrogate gradient as
\begin{align}
    \label{eq:QIFSG}
    \frac{\partial o(t)}{\partial u(t)} \approx 
    \begin{cases} 
      1, & \text{if } \mu_u - \sigma_u \leq u(t) \leq \mu_u + \sigma_u \\
      0, & \text{otherwise}.
   \end{cases}
\end{align}

\section{Experiments}
\label{sec:experiments}

To evaluate the performance of our proposed discretization of the QIF neuron model, we compare our work to state-of-the-art methods on CIFAR-10 and CIFAR-100~\citep{CIFAR10_CIFAR100} with ResNet-19~\citep{HanleZheng-2020}, ImageNet~\citep{ImageNet} with ResNet-34~\citep{HanleZheng-2020}, and CIFAR-10 DVS~\citep{CIFAR10DVS} with ResNet-19 and VGGSNN~\citep{ShikuangDeng-2022}. Additionally, we discuss the energy consumption of models trained with our discretization and analyze its overhead compared to the LIF neuron model.

\subsection{Comparison to Recent Works}
\label{subsection:QIFvsSOTA}

 We adopt the same training setup as recent state-of-the-art works, such as \citet{ShuangLian-2024, YufeiGuo-2024, YufeiGuo-2023, XingtingYao-2022}, for fair comparison of results. We train each model with and without TET loss \cite{ShikuangDeng-2022} to showcase compatibility with previous techniques and present the mean and standard deviation of three trials. ImageNet results only utilize a single trial due to its significant computational cost. Lastly, we separate results within each table to distinguish between works that directly modify the neuron model and works that improve other aspects of SNNs. Complete details on datasets, augmentations, and training setups can be found in Appendices \ref{appendix:datasets_and_augmentations} and \ref{appendix:training_setup}. 

\begin{table*}[t]
\caption{Summary and comparison of accuracy results on static datasets. The \textbf{Timesteps} column correlates to \textbf{CIFAR-10}/\textbf{CIFAR-100}/\textbf{ImageNet}. All \textbf{CIFAR-10}/\textbf{CIFAR-100} results utilize ResNet-19 while all \textbf{ImageNet} results utilize ResNet-34. Acronyms: Surrogate Gradient (SG).}
\centering
\scriptsize
\resizebox{\linewidth}{!}{
\begin{tabular}{lrcccc}
    \toprule
    \textbf{Work} & \textbf{Method} & \centering \textbf{Timesteps} & \centering \textbf{CIFAR-10} & \centering \textbf{CIFAR-100} & \centering\arraybackslash \textbf{ImageNet} \\
    \midrule
     STBP-tdBN \cite{HanleZheng-2020} & Batch Normalization & \centering 6/6/6 & \centering 93.16\% & \centering 71.12\% & \centering\arraybackslash 63.72\% \\
     TEBN \cite{ChaotengDuan-2022} & Batch Normalization & \centering 4/4/4 & \centering 94.71\% & \centering 76.41\% & \centering\arraybackslash 68.28\% \\
     TET \cite{ShikuangDeng-2022} & Loss Function & \centering 6/6/6 & \centering 94.50\% & \centering 74.72\% & \centering\arraybackslash 68.00\%\\
     IM-Loss \cite{YufeiGuo-2022} & Loss Function + SG & \centering 2/2/6 & \centering 93.85\% & \centering 70.18\% & \centering\arraybackslash 67.43\% \\
     LocalZO + TET \cite{BhaskarMukoty-2023} & Direct Training & \centering 2/2/$\times$ & \centering 95.03\% & \centering 76.36\% & \centering\arraybackslash $\times$ \\
     Surrogate Module \cite{ShikuangDeng-2023} & Hybrid & \centering 4/4/4 & \centering 96.82\% & \centering 79.18\% & \centering\arraybackslash 68.25\% \\
     MPBN \cite{YufeiGuo-2023} & Membrane Normalization & \centering 2/2/4 & \centering 96.47\% & \centering 79.51\% & \centering\arraybackslash 64.71\% \\
     Backpropagation Shortcuts \cite{YufeiGuo-2024-Shortcut} & Direct Training & \centering 2/2/4 & \centering 95.19\% & \centering 77.56\% & \centering\arraybackslash 67.90\% \\
     IMP + LTS \cite{HangchiShen-2024} & Membrane Initialization + Loss Function & \centering $\times$/$\times$/4 & \centering $\times$ & \centering $\times$ & \centering\arraybackslash 68.90\% \\
     \cmidrule{1-6}
     \multirow{1}{*}{PLIF \cite{WeiFang-2021}} & \multirow{1}{*}{Neuron Model} 
           & \centering 8/$\times$/$\times$ & \centering 93.50\% & \centering\arraybackslash $\times$ & \centering\arraybackslash $\times$ \\
    \cmidrule{2-6}
    \multirow{2}{*}{GLIF \cite{XingtingYao-2022}} & \multirow{2}{*}{Neuron Model} 
           & \centering 4/4/4 & \centering 94.85\% & \centering\arraybackslash 77.05\% & \centering\arraybackslash 67.52\% \\
     &  & \centering 2/2/$\times$ & \centering 94.44\% & \centering\arraybackslash 75.48\%  & \centering\arraybackslash $\times$ \\ 
     \cmidrule{2-6}
     \multirow{2}{*}{LSG \cite{ShuangLian-2023}} & \multirow{2}{*}{Neuron Model + SG}  
           & \centering 4/4/$\times$ & \centering 95.17\% & \centering\arraybackslash 76.85\% & \centering\arraybackslash $\times$ \\ 
     &  & \centering 2/2/$\times$ & \centering 94.41\% & \centering\arraybackslash 76.32\% & \centering\arraybackslash $\times$ \\ 
     \cmidrule{2-6}
     IM-LIF \cite{ShuangLian-2024} & Neuron Model + Loss Function & \centering 3/3/$\times$ & \centering 95.29\% & \centering 77.21\% & \centering\arraybackslash $\times$\\
     \cmidrule{2-6}
     Ternary Spike \cite{YufeiGuo-2024} & Neuron Model & \centering 2/2/4 & \centering 95.80\% & \centering 80.20\% & \centering\arraybackslash 70.74\% \\
     \cmidrule{2-6}
     \multirow{2}{*}{\textbf{QIF (Ours)}} & \multirow{2}{*}{\textbf{Neuron Model}} 
           & \centering \textbf{4}/\textbf{4}/\textbf{4} & \centering \textbf{96.49 $\pm$ 0.09\%} & \centering \textbf{80.15 $\pm$ 0.06\%} & \centering\arraybackslash \textbf{70.52\%}\\ 
     &  & \centering \textbf{2}/\textbf{2}/$\times$ & \centering \textbf{96.44 $\pm$ 0.07\%} & \centering \textbf{79.68 $\pm$ 0.09\%} & \centering\arraybackslash $\times$\\
     \cmidrule{2-6}
     \multirow{2}{*}{\textbf{QIF (Ours) + TET}} & \multirow{2}{*}{\textbf{Neuron Model}} 
           & \centering \textbf{4}/\textbf{4}/\textbf{4} & \centering \textbf{96.86 $\pm$ 0.13\%} & \centering \textbf{80.62 $\pm$ 0.24\%} & \centering\arraybackslash \textbf{70.38\%}\\ 
     &  & \centering \textbf{2}/\textbf{2}/$\times$ & \centering \textbf{96.70 $\pm$ 0.08\%} & \centering \textbf{80.31 $\pm$ 0.27\%} & \centering\arraybackslash $\times$\\
    \bottomrule
\end{tabular}
}
\label{table:results}
\end{table*}

\textbf{CIFAR-10:} We first evaluate our method on CIFAR-10 using ResNet-19 with and without TET loss. The previous best-performing work achieves $96.47$\% and $96.82$\% with $2$ and $4$ timesteps, respectively. Without TET loss, our discretized QIF neuron model achieves comparable accuracies of $96.44\%$ and $96.49\%$, respectively. With the inclusion of TET loss, our discretization's performance increases to $96.70\%$ and $96.86\%$, surpassing the previous state-of-the-art. 

\textbf{CIFAR-100:}
Next, we similarly evaluate our method on CIFAR-100 using ResNet-19. The previous state-of-the-art works achieve $80.20$\% and $79.18$\% accuracy with $2$ and $4$ timesteps, respectively. Without TET, our discretized QIF neuron model reaches $79.68$\% and $80.15$\% with $2$ and $4$ timesteps, reaching comparable accuracy to previous work. When utilizing TET loss, we increase our accuracy to $80.31$\% and $80.62$\% with $2$ and $4$ timesteps, surpassing the previous state-of-the-art results. 

\textbf{ImageNet:}
We first exclude the results of \citet{YufeiGuo-2024} from our comparison. Without TET, our method surpasses previous approaches by $1.62\%$, and achieves an improvement of $1.48\%$ with TET. These gains highlight the scalability of our discretization to deeper architectures and more complex datasets. The work of \citet{YufeiGuo-2024} outperforms ours by a margin of $0.22-0.36\%$, due to their use of ternary spikes to mitigate information loss and enhance accuracy. In contrast, our results demonstrate that increasing the complexity of neuron dynamics offers a competitive and biologically plausible alternative for improving performance.

\textbf{CIFAR-10 DVS:}
Compared to publications solely focused on neuron model improvements that utilize VGGSNN, we surpass the previous best-performing work of \citet{ShuangLian-2023} by $3.67$\% and $5.1$\% with and without TET, respectively. When looking at all related works, the work of \citet{ChaotengDuan-2022} 84.90\% and outperforms our method. However, the method of \citet{ChaotengDuan-2022} does not allow for batch normalization parameters to be folded into the preceding layers' weights during inference and therefore introduces considerable performance overhead. With ResNet-19, our QIF discretization outperforms the previous best-performing work of \citet{YufeiGuo-2024} by $0.86$\% without TET. Across both models, TET does not improve model performance.

\begin{table*}[t!]
\caption{Comparison between state-of-the-art techniques and our proposed discretization of the QIF neuron model on CIFAR-10 DVS. Acronyms: Surrogate Gradient (SG)}
\centering
\scriptsize
\resizebox{\linewidth}{!}{
\begin{tabular}{llrrcc}
    \toprule
    \textbf{Dataset} & \textbf{Work} & \textbf{Method} & \textbf{Architecture} & \centering \textbf{Timesteps} & \textbf{Accuracy} \\
    \midrule
    \multirow{12}{*}{CIFAR-10 DVS}
     & STDP-tdBN \cite{HanleZheng-2020} & Batch Normalization & ResNet-19 & 10 & 67.80\% \\
    
     & TEBN \cite{ChaotengDuan-2022} & Batch Normalization & VGGSNN & \centering 10 & \centering\arraybackslash 84.90\% \\
     & TET \cite{ShikuangDeng-2022} & Loss Function &  VGGSNN & \centering 10 & \centering\arraybackslash 77.40\% \\
     & IM-Loss \cite{YufeiGuo-2022} & Loss Function + SG & ResNet-19 & \centering 10 & \centering\arraybackslash 72.60\% \\
     & LocalZO + TET \cite{BhaskarMukoty-2023} & Direct Training & VGGSNN & \centering 10 & \centering\arraybackslash 75.62\% \\  
    \cmidrule{2-6}
     & GLIF \cite{XingtingYao-2022} & Neuron Model & ResNet-19 & \centering 16 & \centering\arraybackslash 78.10\% \\
     & LSG \cite{ShuangLian-2023} & Neuron Model + SG & VGGSNN & \centering 10 & \centering\arraybackslash 77.90\% \\
     & Ternary Spike \cite{YufeiGuo-2024} & Neuron Model & ResNet-19 & \centering 10 & \centering\arraybackslash 79.84\% \\
     \cmidrule{3-6}
     & \multirow{2}{*}{\textbf{QIF (Ours)}} & \multirow{2}{*}{\textbf{Neuron Model}} & \textbf{VGGSNN} & \centering \textbf{10} & \centering\arraybackslash \textbf{83.00 $\pm$ 0.50\%} \\
     & & & \textbf{ResNet-19} & \centering \textbf{10} & \centering\arraybackslash \textbf{80.70 $\pm$ 0.10\%} \\
     \cmidrule{3-6}
     & \multirow{2}{*}{\textbf{QIF (Ours) + TET}} & \multirow{2}{*}{\textbf{Neuron Model}} & \textbf{VGGSNN } & \centering \textbf{10} & \centering\arraybackslash \textbf{81.57 $\pm$ 0.92\%} \\
     & & & \textbf{ResNet-19} & \centering \textbf{10} & \centering\arraybackslash \textbf{75.10 $\pm$ 1.50\%} \\
    \bottomrule
\end{tabular}
}
\label{table:results_cont}
\end{table*}

\subsection{Energy Consumption}

To calculate the energy consumption of SNNs, we adopt the approach of \citet{QiaoyiSu-2023}, where they approximate it as $E_\text{SNN} \approx \sum_i E_i$. $E_i$ is the energy consumption of layer $i$ and is defined as 
\begin{align}
    \label{eq:SNNEnergyConsumption}
    E_i = T \cdot (fr \cdot E_\text{AC} \cdot OP_\text{AC} + E_\text{MAC} \cdot OP_\text{MAC})
\end{align}
where $T$ is the number of timesteps, $fr$ is the firing rate of layer $i$, $E_\text{AC}$ and $E_\text{MAC}$ are the energy consumption of accumulate (AC) and multiply-and-accumulate (MAC) operations respectively, and $OP_\text{AC}$ and $OP_\text{MAC}$ are the number of AC and MAC operations in layer $i$. These operations are assumed to take place with $32$-bit floating point values on $45$nm technology where $E_\text{MAC} = 4.6pJ$ and $E_\text{AC} = 0.9pJ$, as done by~\citet{ShuangLian-2024}. Table~\ref{table:energy_consumption} breaks down the energy cost per model and dataset using~\eqref{eq:SNNEnergyConsumption}. Our discretized QIF neuron model has more MAC operations than the LIF neuron model. Since MACs aren't dependent on the spike rate, we calculate their overhead and present our results in Table~\ref{table:energy_consumption}. Across all models and experiments, our discretization only introduces a small energy consumption overhead between $0.94\% - 3.23\%$ compared to the LIF neuron model. 

\begin{table*}[h]
    \caption{Estimated energy consumption per inference of our proposed discretization of the QIF neuron model across our examined datasets and model architectures.}
    \centering
    \scriptsize
    \resizebox{\linewidth}{!}{
    \begin{tabular}{llcrrrr}
        \toprule
        \textbf{Architecture} & \textbf{Dataset} & \textbf{Timesteps} & \textbf{MAC Operations} & \textbf{AC Operations} & \textbf{Energy Consumption} & \textbf{QIF Overhead}\\
        \midrule
        \multirow{2}{*}{\textbf{ResNet-19}} & CIFAR-10 & 2 & \multirow{2}{*}{$4.98 \times 10^6$} & \multirow{2}{*}{$2.21 \times 10^9$} & $0.633 mJ$ & $0.008 mJ$ / $1.20\%$ \\
        & CIFAR-10 & 4 &  &  & $1.253 mJ$ & $0.015 mJ$ / $1.21\%$ \\
        \midrule
        \multirow{2}{*}{\textbf{ResNet-19}} & CIFAR-100 & 2 & \multirow{2}{*}{$4.98 \times 10^6$} & \multirow{2}{*}{$2.21 \times 10^9$} & $0.809 mJ$ & $0.008 mJ$ / $0.94\%$ \\
        & CIFAR-100 & 4 &  &  & $1.578 mJ$ & $0.015 mJ$ / $0.96\%$ \\
        \midrule
        \textbf{ResNet-34} & ImageNet & 4 & $1.25 \times 10^8$ & $3.66 \times10^9$ & $6.402 mJ$ & $0.069 mJ$ / $1.07\%$ \\
        \midrule
        \textbf{VGGSNN} & CIFAR-10 DVS & 10 & $4.50 \times 10^6$ & $1.36 \times 10^9$ & $1.312 mJ$ & $0.042 mJ$ / $3.23\%$  \\
        \textbf{ResNet-19} & CIFAR-10 DVS & 10 & $9.88 \times 10^6$ & $4.98 \times 10^9$ & $11.998 mJ$ & $0.177 mJ$ / $1.48\%$  \\
        \bottomrule
    \end{tabular}
    }
    \label{table:energy_consumption}
\end{table*}

\section{Limitations}

One limitation of our QIF discretization is its reliance on tdBN~\eqref{eq:tdBN} for surrogate gradient calculation, which hinders event-driven backpropagation~\citep{TimoWunderlich-2021} since each tdBN layer performs normalization across both spatial and temporal dimensions. Nevertheless, this does not impede event-driven inference, as tdBN parameters can be merged into adjacent layers~\citep{HanleZheng-2020}. Another limitation lies in the increased computational cost relative to the LIF model~\eqref{eq:LIF}. Our formulation and parameter choice introduces an additional multiplication and addition per update. Although this adds negligible energy overhead, its impact on computational performance within neuromorphic systems remains uncertain due to limited commercial availability. Finally, the additional hyperparameters of the QIF neuron may increase the difficulty of training deep SNNs in other settings.

\section{Conclusion}

This work introduces a novel discretization of the QIF neuron, enabling scalable, high-performance deep SNNs with richer dynamics than LIF neurons. Our discretization captures subthreshold oscillations and input sensitivity while supporting stable training via analytically derived surrogate gradients. With minimal energy overhead, our experiments demonstrate consistent improvement over the current state-of-the-art across CIFAR-10, CIFAR-100, and CIFAR-10 DVS, while showcasing competitive performance on ImageNet. These results highlight our discretization of the QIF neuron as a promising alternative to LIF neurons for accurate and energy-efficient deep SNNs.

\bibliographystyle{plainnat}
\bibliography{refs}

\begin{thebibliography}{39}
\providecommand{\natexlab}[1]{#1}
\providecommand{\url}[1]{\texttt{#1}}
\expandafter\ifx\csname urlstyle\endcsname\relax
  \providecommand{\doi}[1]{doi: #1}\else
  \providecommand{\doi}{doi: \begingroup \urlstyle{rm}\Url}\fi

\bibitem[Akopyan et~al.(2015)Akopyan, Sawada, Cassidy, Alvarez-Icaza, Arthur, Merolla, Imam, Nakamura, Datta, Nam, Taba, Beakes, Brezzo, Kuang, Manohar, Risk, Jackson, and Modha]{FilippAkopyan-2015}
Filipp Akopyan, Jun Sawada, Andrew Cassidy, Rodrigo Alvarez-Icaza, John Arthur, Paul Merolla, Nabil Imam, Yutaka Nakamura, Pallab Datta, Gi-Joon Nam, Brian Taba, Michael Beakes, Bernard Brezzo, Jente~B. Kuang, Rajit Manohar, William~P. Risk, Bryan Jackson, and Dharmendra~S. Modha.
\newblock Truenorth: Design and tool flow of a 65 mw 1 million neuron programmable neurosynaptic chip.
\newblock \emph{IEEE Transactions on Computer-Aided Design of Integrated Circuits and Systems}, 34\penalty0 (10):\penalty0 1537--1557, 2015.
\newblock \doi{10.1109/TCAD.2015.2474396}.

\bibitem[Cao et~al.(2015)Cao, Chen, and Khosla]{YongqiangCao-2015}
Yongqiang Cao, Yang Chen, and Deepak Khosla.
\newblock Spiking deep convolutional neural networks for energy-efficient object recognition.
\newblock \emph{International Journal of Computer Vision}, 113\penalty0 (1):\penalty0 54--66, May 2015.
\newblock ISSN 1573-1405.
\newblock \doi{10.1007/s11263-014-0788-3}.
\newblock URL \url{https://doi.org/10.1007/s11263-014-0788-3}.

\bibitem[Chai et~al.(2021)Chai, Zeng, Li, and Ngai]{JunyiChai-2021}
Junyi Chai, Hao Zeng, Anming Li, and Eric~W.T. Ngai.
\newblock Deep learning in computer vision: A critical review of emerging techniques and application scenarios.
\newblock \emph{Machine Learning with Applications}, 6:\penalty0 100134, 2021.
\newblock ISSN 2666-8270.
\newblock \doi{https://doi.org/10.1016/j.mlwa.2021.100134}.
\newblock URL \url{https://www.sciencedirect.com/science/article/pii/S2666827021000670}.

\bibitem[Cubuk et~al.(2019)Cubuk, Zoph, Mané, Vasudevan, and Le]{EkinCubuk-2019}
Ekin~D. Cubuk, Barret Zoph, Dandelion Mané, Vijay Vasudevan, and Quoc~V. Le.
\newblock Autoaugment: Learning augmentation strategies from data.
\newblock In \emph{2019 IEEE/CVF Conference on Computer Vision and Pattern Recognition (CVPR)}, pages 113--123, 2019.
\newblock \doi{10.1109/CVPR.2019.00020}.

\bibitem[Deng et~al.(2009)Deng, Dong, Socher, Li, Li, and Fei-Fei]{ImageNet}
Jia Deng, Wei Dong, Richard Socher, Li-Jia Li, Kai Li, and Li~Fei-Fei.
\newblock Imagenet: A large-scale hierarchical image database.
\newblock In \emph{2009 IEEE Conference on Computer Vision and Pattern Recognition}, pages 248--255, 2009.
\newblock \doi{10.1109/CVPR.2009.5206848}.

\bibitem[Deng et~al.(2022)Deng, Li, Zhang, and Gu]{ShikuangDeng-2022}
Shikuang Deng, Yuhang Li, Shanghang Zhang, and Shi Gu.
\newblock Temporal efficient training of spiking neural network via gradient re-weighting.
\newblock In \emph{International Conference on Learning Representations}, 2022.
\newblock URL \url{https://openreview.net/forum?id=_XNtisL32jv}.

\bibitem[Deng et~al.(2023)Deng, Lin, Li, and Gu]{ShikuangDeng-2023}
Shikuang Deng, Hao Lin, Yuhang Li, and Shi Gu.
\newblock Surrogate module learning: reduce the gradient error accumulation in training spiking neural networks.
\newblock In \emph{Proceedings of the 40th International Conference on Machine Learning}, ICML'23. JMLR.org, 2023.

\bibitem[Duan et~al.(2022)Duan, Ding, Chen, Yu, and Huang]{ChaotengDuan-2022}
Chaoteng Duan, Jianhao Ding, Shiyan Chen, Zhaofei Yu, and Tiejun Huang.
\newblock Temporal effective batch normalization in spiking neural networks.
\newblock \emph{Advances in Neural Information Processing Systems}, 35:\penalty0 34377–34390, December 2022.

\bibitem[Fang et~al.(2021)Fang, Yu, Chen, Masquelier, Huang, and Tian]{WeiFang-2021}
Wei Fang, Zhaofei Yu, Yanqi Chen, Timothée Masquelier, Tiejun Huang, and Yonghong Tian.
\newblock Incorporating learnable membrane time constant to enhance learning of spiking neural networks.
\newblock In \emph{2021 IEEE/CVF International Conference on Computer Vision (ICCV)}, pages 2641--2651, 2021.
\newblock \doi{10.1109/ICCV48922.2021.00266}.

\bibitem[Gerstner et~al.(2014)Gerstner, Kistler, Naud, and Paninski]{WulframGerstner-2014}
Wulfram Gerstner, Werner~M. Kistler, Richard Naud, and Liam Paninski.
\newblock \emph{Neuronal Dynamics: From Single Neurons to Networks and Models of Cognition}.
\newblock Cambridge University Press, 2014.

\bibitem[Guo et~al.(2022)Guo, Chen, Zhang, Liu, Wang, Huang, and Ma]{YufeiGuo-2022}
Yufei Guo, Yuanpei Chen, Liwen Zhang, Xiaode Liu, Yinglei Wang, Xuhui Huang, and Zhe Ma.
\newblock Im-loss: Information maximization loss for spiking neural networks.
\newblock In S.~Koyejo, S.~Mohamed, A.~Agarwal, D.~Belgrave, K.~Cho, and A.~Oh, editors, \emph{Advances in Neural Information Processing Systems}, volume~35, pages 156--166. Curran Associates, Inc., 2022.
\newblock URL \url{https://proceedings.neurips.cc/paper_files/paper/2022/file/010c5ba0cafc743fece8be02e7adb8dd-Paper-Conference.pdf}.

\bibitem[Guo et~al.(2023)Guo, Zhang, Chen, Peng, Liu, Zhang, Huang, and Ma]{YufeiGuo-2023}
Yufei Guo, Yuhan Zhang, Yuanpei Chen, Weihang Peng, Xiaode Liu, Liwen Zhang, Xuhui Huang, and Zhe Ma.
\newblock Membrane potential batch normalization for spiking neural networks.
\newblock In \emph{2023 IEEE/CVF International Conference on Computer Vision (ICCV)}, pages 19363--19373, 2023.
\newblock \doi{10.1109/ICCV51070.2023.01779}.

\bibitem[Guo et~al.(2024{\natexlab{a}})Guo, Chen, Hao, Peng, Jie, Zhang, Liu, and Ma]{YufeiGuo-2024-Shortcut}
Yufei Guo, Yuanpei Chen, Zecheng Hao, Weihang Peng, Zhou Jie, Yuhan Zhang, Xiaode Liu, and Zhe Ma.
\newblock Take a shortcut back: Mitigating the gradient vanishing for training spiking neural networks.
\newblock In A.~Globerson, L.~Mackey, D.~Belgrave, A.~Fan, U.~Paquet, J.~Tomczak, and C.~Zhang, editors, \emph{Advances in Neural Information Processing Systems}, volume~37, pages 24849--24867. Curran Associates, Inc., 2024{\natexlab{a}}.
\newblock URL \url{https://proceedings.neurips.cc/paper_files/paper/2024/file/2c37c5bcef24b9541550261dcd63261b-Paper-Conference.pdf}.

\bibitem[Guo et~al.(2024{\natexlab{b}})Guo, Chen, Liu, Peng, Zhang, Huang, and Ma]{YufeiGuo-2024}
Yufei Guo, Yuanpei Chen, Xiaode Liu, Weihang Peng, Yuhan Zhang, Xuhui Huang, and Zhe Ma.
\newblock Ternary spike: learning ternary spikes for spiking neural networks.
\newblock In \emph{Proceedings of the Thirty-Eighth AAAI Conference on Artificial Intelligence and Thirty-Sixth Conference on Innovative Applications of Artificial Intelligence and Fourteenth Symposium on Educational Advances in Artificial Intelligence}, AAAI'24/IAAI'24/EAAI'24. AAAI Press, 2024{\natexlab{b}}.
\newblock ISBN 978-1-57735-887-9.
\newblock \doi{10.1609/aaai.v38i11.29114}.
\newblock URL \url{https://doi.org/10.1609/aaai.v38i11.29114}.

\bibitem[Intel(2021)]{IntelLoihi2}
Intel.
\newblock Taking neuromorphic computing to the next level with loihi 2 technology brief.
\newblock \url{https://www.intel.com/content/www/us/en/research/neuromorphic-computing-loihi-2-technology-brief.html}, 2021.
\newblock Accessed: 03-19-2024.

\bibitem[Ioffe and Szegedy(2015)]{SergeyIoffe-2015}
Sergey Ioffe and Christian Szegedy.
\newblock Batch normalization: Accelerating deep network training by reducing internal covariate shift.
\newblock In Francis Bach and David Blei, editors, \emph{Proceedings of the 32nd International Conference on Machine Learning}, volume~37 of \emph{Proceedings of Machine Learning Research}, pages 448--456, Lille, France, 07--09 Jul 2015. PMLR.
\newblock URL \url{https://proceedings.mlr.press/v37/ioffe15.html}.

\bibitem[jhu Lee et~al.(2021)jhu Lee, On, Xiao, Proietti, and Yoo]{YunJhuLee-2021}
Yun jhu Lee, Mehmet~Berkay On, Xian Xiao, Roberto Proietti, and S.~J.~Ben Yoo.
\newblock Izhikevich-inspired optoelectronic neurons with excitatory and inhibitory inputs for energy-efficient photonic spiking neural networks, 2021.
\newblock URL \url{https://arxiv.org/abs/2105.02809}.

\bibitem[Khan et~al.(2023)Khan, Daud, Khan, Muhammad, and Haq]{WahabKhan-2023}
Wahab Khan, Ali Daud, Khairullah Khan, Shakoor Muhammad, and Rafiul Haq.
\newblock Exploring the frontiers of deep learning and natural language processing: A comprehensive overview of key challenges and emerging trends.
\newblock \emph{Natural Language Processing Journal}, 4:\penalty0 100026, 2023.
\newblock ISSN 2949-7191.
\newblock \doi{https://doi.org/10.1016/j.nlp.2023.100026}.
\newblock URL \url{https://www.sciencedirect.com/science/article/pii/S2949719123000237}.

\bibitem[Krizhevsky(2009)]{CIFAR10_CIFAR100}
Alex Krizhevsky.
\newblock Learning multiple layers of features from tiny images.
\newblock \emph{""}, pages 32--33, 2009.
\newblock URL \url{https://www.cs.toronto.edu/~kriz/learning-features-2009-TR.pdf}.

\bibitem[Lampl and Yarom(1993)]{IlanLampl-1993}
I.~Lampl and Y.~Yarom.
\newblock Subthreshold oscillations of the membrane potential: a functional synchronizing and timing device.
\newblock \emph{Journal of Neurophysiology}, 70\penalty0 (5):\penalty0 2181--2186, 1993.
\newblock \doi{10.1152/jn.1993.70.5.2181}.
\newblock URL \url{https://doi.org/10.1152/jn.1993.70.5.2181}.
\newblock PMID: 8294979.

\bibitem[Lenz et~al.(2023)Lenz, Orchard, and Sheik]{GregorLenz-2023}
Gregor Lenz, Garrick Orchard, and Sadique Sheik.
\newblock Ultra-low-power image classification on neuromorphic hardware, 2023.

\bibitem[Li et~al.(2017)Li, Liu, Ji, Li, and Shi]{CIFAR10DVS}
Hongmin Li, Hanchao Liu, Xiangyang Ji, Guoqi Li, and Luping Shi.
\newblock Cifar10-dvs: An event-stream dataset for object classification.
\newblock \emph{Frontiers in Neuroscience}, 11, 2017.
\newblock ISSN 1662-453X.
\newblock \doi{10.3389/fnins.2017.00309}.
\newblock URL \url{https://www.frontiersin.org/journals/neuroscience/articles/10.3389/fnins.2017.00309}.

\bibitem[Li et~al.(2021)Li, Guo, Zhang, Deng, Hai, and Gu]{YuhangLi-2021}
Yuhang Li, Yufei Guo, Shanghang Zhang, Shikuang Deng, Yongqing Hai, and Shi Gu.
\newblock Differentiable spike: Rethinking gradient-descent for training spiking neural networks.
\newblock In \emph{Advances in Neural Information Processing Systems}, volume~34, page 23426–23439. Curran Associates, Inc., 2021.
\newblock URL \url{proceedings.neurips.cc/paper_files/paper/2021/hash/c4ca4238a0b923820dcc509a6f75849b-Abstract.html}.

\bibitem[Lian et~al.(2023)Lian, Shen, Liu, Wang, Yan, and Tang]{ShuangLian-2023}
Shuang Lian, Jiangrong Shen, Qianhui Liu, Ziming Wang, Rui Yan, and Huajin Tang.
\newblock Learnable surrogate gradient for direct training spiking neural networks.
\newblock In \emph{Proceedings of the Thirty-Second International Joint Conference on Artificial Intelligence}, page 3002–3010, Macau, SAR China, August 2023. International Joint Conferences on Artificial Intelligence Organization.
\newblock ISBN 978-1-956792-03-4.
\newblock \doi{10.24963/ijcai.2023/335}.
\newblock URL \url{https://www.ijcai.org/proceedings/2023/335}.

\bibitem[Lian et~al.(2024)Lian, Shen, Wang, and Tang]{ShuangLian-2024}
Shuang Lian, Jiangrong Shen, Ziming Wang, and Huajin Tang.
\newblock Im-lif: Improved neuronal dynamics with attention mechanism for direct training deep spiking neural network.
\newblock \emph{IEEE Transactions on Emerging Topics in Computational Intelligence}, 8\penalty0 (2):\penalty0 2075--2085, 2024.
\newblock \doi{10.1109/TETCI.2024.3359539}.

\bibitem[Mukhoty et~al.(2023)Mukhoty, Bojkovic, de~Vazelhes, Zhao, Masi, Xiong, and Gu]{BhaskarMukoty-2023}
Bhaskar Mukhoty, Velibor Bojkovic, William de~Vazelhes, Xiaohan Zhao, Giulia~De Masi, Huan Xiong, and Bin Gu.
\newblock Direct training of {SNN} using local zeroth order method.
\newblock In \emph{Thirty-seventh Conference on Neural Information Processing Systems}, 2023.
\newblock URL \url{https://openreview.net/forum?id=eTF3VDH2b6}.

\bibitem[Nirenberg and Meister(1997)]{Nirenberg1997}
Sheila Nirenberg and Markus Meister.
\newblock The light response of retinal ganglion cells is truncated by a displaced amacrine circuit.
\newblock \emph{Neuron}, 18\penalty0 (4):\penalty0 637--650, 1997.
\newblock ISSN 0896-6273.
\newblock \doi{https://doi.org/10.1016/S0896-6273(00)80304-9}.
\newblock URL \url{https://www.sciencedirect.com/science/article/pii/S0896627300803049}.

\bibitem[Rathi and Roy(2023)]{NitinRathi-2023}
Nitin Rathi and Kaushik Roy.
\newblock Diet-snn: A low-latency spiking neural network with direct input encoding and leakage and threshold optimization.
\newblock \emph{IEEE Transactions on Neural Networks and Learning Systems}, 34\penalty0 (6):\penalty0 3174--3182, 2023.
\newblock \doi{10.1109/TNNLS.2021.3111897}.

\bibitem[Shen et~al.(2024)Shen, Zheng, Wang, and Pan]{HangchiShen-2024}
Hangchi Shen, Qian Zheng, Huamin Wang, and Gang Pan.
\newblock Rethinking the membrane dynamics and optimization objectives of spiking neural networks.
\newblock In A.~Globerson, L.~Mackey, D.~Belgrave, A.~Fan, U.~Paquet, J.~Tomczak, and C.~Zhang, editors, \emph{Advances in Neural Information Processing Systems}, volume~37, pages 92697--92720. Curran Associates, Inc., 2024.
\newblock URL \url{https://proceedings.neurips.cc/paper_files/paper/2024/file/a867a94a427bacbe3f4de16c7ac10ba8-Paper-Conference.pdf}.

\bibitem[Su et~al.(2023)Su, Chou, Hu, Li, Mei, Zhang, and Li]{QiaoyiSu-2023}
Qiaoyi Su, Yuhong Chou, Yifan Hu, Jianing Li, Shijie Mei, Ziyang Zhang, and Guoqi Li.
\newblock Deep directly-trained spiking neural networks for object detection.
\newblock In \emph{2023 IEEE/CVF International Conference on Computer Vision (ICCV)}, page 6532–6542, Paris, France, October 2023. IEEE.
\newblock ISBN 9798350307184.
\newblock \doi{10.1109/ICCV51070.2023.00603}.
\newblock URL \url{https://ieeexplore.ieee.org/document/10376840/}.

\bibitem[Tsai et~al.(2017)Tsai, Barch, Cassidy, DeBole, Andreopoulos, Jackson, Flickner, Arthur, Modha, Sampson, and Narayanan]{Wei-YuTasi-2017}
Wei-Yu Tsai, Davis~R. Barch, Andrew~S. Cassidy, Michael~V. DeBole, Alexander Andreopoulos, Bryan~L. Jackson, Myron~D. Flickner, John~V. Arthur, Dharmendra~S. Modha, John Sampson, and Vijaykrishnan Narayanan.
\newblock Always-on speech recognition using truenorth, a reconfigurable, neurosynaptic processor.
\newblock \emph{IEEE Transactions on Computers}, 66\penalty0 (6):\penalty0 996--1007, 2017.
\newblock \doi{10.1109/TC.2016.2630683}.

\bibitem[Wu et~al.(2021)Wu, Yeh, White, Wang, Yeh, Hsieh, Liu, Tang, and Lo]{WenChiehWu-2021}
Wen-Chieh Wu, Chen-Fu Yeh, Alexander~James White, Cheng-Te Wang, Zuo-Wei Yeh, Chih-Cheng Hsieh, Ren-Shuo Liu, Kea-Tiong Tang, and Chung-Chuan Lo.
\newblock Integer quadratic integrate-and-fire (iqif): A neuron model for digital neuromorphic systems.
\newblock In \emph{2021 IEEE 3rd International Conference on Artificial Intelligence Circuits and Systems (AICAS)}, pages 1--4, 2021.
\newblock \doi{10.1109/AICAS51828.2021.9458572}.

\bibitem[Wu et~al.(2018)Wu, Deng, Li, Zhu, and Shi]{YujieWu-2018}
Yujie Wu, Lei Deng, Guoqi Li, Jun Zhu, and Luping Shi.
\newblock Spatio-temporal backpropagation for training high-performance spiking neural networks.
\newblock \emph{Frontiers in Neuroscience}, 12, 2018.
\newblock ISSN 1662-453X.
\newblock \doi{10.3389/fnins.2018.00331}.
\newblock URL \url{https://www.frontiersin.org/journals/neuroscience/articles/10.3389/fnins.2018.00331}.

\bibitem[Wunderlich and Pehle(2021)]{TimoWunderlich-2021}
Timo~C. Wunderlich and Christian Pehle.
\newblock Event-based backpropagation can compute exact gradients for spiking neural networks.
\newblock \emph{Scientific Reports}, 11\penalty0 (1):\penalty0 12829, Jun 2021.
\newblock ISSN 2045-2322.
\newblock \doi{10.1038/s41598-021-91786-z}.
\newblock URL \url{https://doi.org/10.1038/s41598-021-91786-z}.

\bibitem[Yamazaki et~al.(2022)Yamazaki, Vo-Ho, Bulsara, and Le]{KashuYamazaki-2022}
Kashu Yamazaki, Viet-Khoa Vo-Ho, Darshan Bulsara, and Ngan Le.
\newblock Spiking neural networks and their applications: A review.
\newblock \emph{Brain Sci}, 12\penalty0 (7), June 2022.

\bibitem[Yao et~al.(2022)Yao, Li, Mo, and Cheng]{XingtingYao-2022}
Xingting Yao, Fanrong Li, Zitao Mo, and Jian Cheng.
\newblock {GLIF}: A unified gated leaky integrate-and-fire neuron for spiking neural networks.
\newblock In Alice~H. Oh, Alekh Agarwal, Danielle Belgrave, and Kyunghyun Cho, editors, \emph{Advances in Neural Information Processing Systems}, 2022.
\newblock URL \url{https://openreview.net/forum?id=UmFSx2c4ubT}.

\bibitem[Ye(2021)]{WeijieYe-2021}
Weijie Ye.
\newblock Dynamics of a {Large-Scale} spiking neural network with quadratic {Integrate-and-Fire} neurons.
\newblock \emph{Neural Plast}, 2021:\penalty0 6623926, February 2021.

\bibitem[Yi et~al.(2023)Yi, Lian, Liu, Zhu, Liang, and Liu]{ZexiangYi-2023}
Zexiang Yi, Jing Lian, Qidong Liu, Hegui Zhu, Dong Liang, and Jizhao Liu.
\newblock Learning rules in spiking neural networks: A survey.
\newblock \emph{Neurocomputing}, 531:\penalty0 163--179, 2023.
\newblock ISSN 0925-2312.
\newblock \doi{https://doi.org/10.1016/j.neucom.2023.02.026}.
\newblock URL \url{https://www.sciencedirect.com/science/article/pii/S0925231223001662}.

\bibitem[Zheng et~al.(2021)Zheng, Wu, Deng, Hu, and Li]{HanleZheng-2020}
Hanle Zheng, Yujie Wu, Lei Deng, Yifan Hu, and Guoqi Li.
\newblock Going deeper with directly-trained larger spiking neural networks.
\newblock \emph{Proceedings of the AAAI Conference on Artificial Intelligence}, 35\penalty0 (12):\penalty0 11062--11070, May 2021.
\newblock \doi{10.1609/aaai.v35i12.17320}.
\newblock URL \url{https://ojs.aaai.org/index.php/AAAI/article/view/17320}.

\end{thebibliography}


\appendix

\newpage
\section{Datasets and Augmentations}
\label{appendix:datasets_and_augmentations}

\subsection{CIFAR-10}

\textbf{CIFAR-10} \citep{CIFAR10_CIFAR100} is a widely used dataset for traditional ANN and SNN models. It consists of $60,000, 32\times32$ colored images consisting of $10$ classes, with some examples being airplanes, automobiles, cats, and horses. There are $6,000$ images per class. The dataset is split into standardized training and testing sets with $50,000$ and $10,000$ images, respectively. When training, we perform the following dataset augmentations: Random cropping after adding $4$ pixels of zero padding to the outside of the image, random horizontal flipping, cutout, and normalization of each image by the mean and standard deviation of the training set. 

\subsection{CIFAR-100}

\textbf{CIFAR-100} \citep{CIFAR10_CIFAR100} is also a widely used dataset for traditional ANN and SNN models. It contains the same number of images and shapes as CIFAR-10. However, CIFAR-100 contains $100$ unique classes instead of $10$, with several examples being bed, rocket, apples, and otter. There are only $600$ images per class, with each class evenly split between standardized training and testing sets the same size as CIFAR-10. We use the same augmentations as with CIFAR-10 in addition to AutoAugment \citep{EkinCubuk-2019}.

\subsection{ImageNet}

\textbf{ImageNet} \citep{ImageNet} is one of the most popular and challenging image classification benchmarks. ImageNet contains $1000$ in different classes, with $1.28$ million training images and $50,000$ validation images in various shapes and sizes. During training, we resize the image to $256\times 256$ and take a random crop of size $224 \times 224$, apply random horizontal flipping, cutout, and normalize each image by the mean and standard deviation of the training set. 

\subsection{CIFAR-10 DVS}

\textbf{CIFAR-10 DVS} \citep{CIFAR10DVS} uses a subset of CIFAR-10 with $1,000$ images from each class. A Digital Vision Sensor (DVS) converts the CIFAR-10 dataset~\citep{CIFAR10_CIFAR100} from real-valued images into spike-based events. These events are accumulated for a set number of timesteps to generate frames of spiking activity with a spatial resolution of $128\times128$.
Using this dataset, we accumulate events into $10$ frames and resize them to $48\times48$. Additionally, we apply random horizontal flipping and randomly rotate the image up to $\pm 10^\circ$.

\newpage
\section{Training Setup}
\label{appendix:training_setup}

We use the following model architectures: ResNet-19 \citet{HanleZheng-2020} and VGGSNN \citet{ShikuangDeng-2022}. Complete details of the training setup and hyperparameters used for each dataset and model are presented in Table \ref{table:hyperparameters}. We use a cosine decay learning rate scheduler to gradually decrease the learning rate to 0 for all models. On ImageNet, we include a learning rate warm-up for the first five epochs. We set $\beta_1 = 0.9$, $\beta_2 = 0.999$ when using Adam. We do not apply weight decay to batch normalization parameters. We use the following parameters for all models trained with our proposed QIF discretization: $u_{th} = 0.5$, $u_1 = 0$, $u_2 = 0.5$, and $a = 0.25$. 

All CIFAR-10, CIFAR-100, and CIFAR-10 DVS experiments utilized a single Nvidia A100 GPU and 32GB of RAM. Our ImageNet experiments utilized four Nvidia A100 GPUs and 256 GB of RAM. CIFAR-10/CIFAR-100 experiments with two timesteps took just over five hours to complete, CIFAR-10/CIFAR-100 experiments with four timesteps took just over 11 hours to complete, experiments with CIFAR-10 DVS using VGGSNN and ResNet-19 took just under an hour and just under three hours to complete, respectively. Our ImageNet experiments each took just over six days to complete. Preliminary and failed experiments did not require additional computing resources.

\begin{table*}[ht]
\caption{Training hyperparameters for each model.}
\centering
\footnotesize
\begin{tabular}{lrrrr}
    \toprule
    \textbf{Parameters} & \textbf{CIFAR-10} & \textbf{CIFAR-100} & \textbf{ImageNet} & \textbf{CIFAR-10 DVS} \\
    \midrule
    Model & ResNet-19 & ResNet-19 & ResNet-34 & VGGSNN / ResNet-19 \\
    Optimizer & SGD & SGD & SGD & Adam \\
    Learning Rate & $0.1$ & $0.1$ & $0.15$ & $1e-3$  \\
    Weight Decay & $1e-4$ & $1e-4$ & $5e-5$ & $5e-4$  \\
    Momentum & $0.9$ & $0.9$ & $0.9$ & $\times$ \\
    Epochs & $400$ & $400$ & $350$ & $100$ \\
    Batch Size & $128$ & $128$ & $512$ & $128$ \\
    Timesteps & $2, 4$ & $2, 4$ & $4$ & $10$ \\
    Dropout & $\times$ & $\times$ & $\times$ & $60$\% \\
    $\lambda$ (TET Only) & $0.05$ & $0.05$ & $0.001$ & $0.001$ \\
    \bottomrule
\end{tabular}
\label{table:hyperparameters}
\end{table*}

\newpage

\newpage
\section{Discretized Quadratic Integrate-and-Fire Neuron Model}
\label{appendix:discretization}

\subsection{Discretization}

The QIF neuron model is defined as
\begin{align}
    \label{eq:appendix_QIFODE}
    \tau \frac{du}{dt} = a(u - u_{r})(u - u_c) + RI,
\end{align}
where $\tau$ is a membrane time constant, $u$ is the voltage, $a$ is a sharpness parameter, $u_{r}$ is the resting voltage, $u_c$ is the critical spiking threshold, $R$ is a resistance, and $I$ is pre-synaptic input. We use Euler's method to discretize~\eqref{eq:appendix_QIFODE}, as done by \citet{YujieWu-2018}. First, we replace the derivative with the following discrete-time derivative
\begin{align}
    \frac{du}{dt} \approx \frac{u(t+1) - u(t)}{\Delta t}.
\end{align}
Substituting this into~\eqref{eq:appendix_QIFODE}, we have
\begin{align}
    \tau \frac{u(t+1) - u(t)}{\Delta t} = a(u(t) - u_{r})(u(t) - u_c) + RI(t),
\end{align}
with $t$ being some discrete timestep and $\Delta t$ being some small step size. Then, solving for $u(t+1)$ gives us
\begin{align}
\label{eq:appendix_QIF_discrete_1}
    u(t+1) = u(t) + \frac{\Delta t}{\tau}[a(u(t) - u_{r})(u(t) - u_c) + RI(t)].
\end{align}
Next, by incorporating $\frac{\Delta t}{\tau}$ into $a$ and $R\frac{\Delta t}{\tau}$ into the learnable weights used to calculate $I(t)$ we can simplify~\eqref{eq:appendix_QIF_discrete_1} to
\begin{align}
    \label{eq:appendx_badQIF}
    u(t+1) = u(t) + a(u(t) - u_{r})(u(t) - u_c) + I(t).
\end{align}
By expanding the terms on the right-hand side of the equation above, one can write 
\begin{align} 
    u(t+1) = \left[a u(t)^2 - a\left(u_r + u_c - \frac{1}{a}\right) u(t) + au_r u_c\right] + I(t), 
\end{align} 
and thus, by rewriting the expression between brackets, the equation
\begin{align} 
\label{eq:appendix_QIF}
    u(t+1) = a(u(t) - u_1)(u(t) - u_2) + I(t), 
\end{align} 
is obtained, where 
\begin{equation}
\label{eq:appendix_QIF_param_calc}
\begin{aligned} 
    u_{1,2} &= \frac{1}{2} \left[(u_r + u_c) - \frac{1}{a}\right] \mp \frac{1}{2} \sqrt{\Delta}\\
    \Delta &= (u_c - u_r)^2 - \frac{2(u_r + u_c)}{a} + \frac{1}{a^2}.
\end{aligned} 
\end{equation}

\subsection{Fixed points and stability}

By evaluating our proposed discretization of the QIF model under special conditions, a better understanding of its dynamics can be obtained. First, consider~\eqref{eq:appendix_QIF} with null input. The minimum value that $u(t+1)$ can assume is calculated as the minimum value of the parabola
\begin{equation}
    y = ax^2 - a(u_1+u_2)x + au_1u_2.
\end{equation}
So,
\begin{equation}
\begin{aligned}
    u_\text{min} &= -\frac{a^2(u_1+u_2)^2-4a^2u_1u_2}{4a}\\ 
        &= -\frac{a^2(u_1-u_2)^2}{4a}\\
        &= -\frac{a(u_1-u_2)^2}{4}.\\    
\end{aligned}
\end{equation}
Under the same conditions, the set of points that make $u(t+1)=0$ are given by solving
\begin{equation}
    0 = a(u(t)-u_1)(u(t)-u_2),
\end{equation}
which gives $u(t)=u_1$ or $u(t)=u_2$. Also, with null input, the set of fixed points is obtained by making $u(t+1)=u(t)$ in~\eqref{eq:appendx_badQIF}. That gives
\begin{equation}
    0 = a(u(t)-u_r)(u(t)-u_c).
\end{equation}
Therefore, $u_c$ and $u_r$ are fixed points of both~\eqref{eq:appendx_badQIF} and~\eqref{eq:appendix_QIF}. The stability of such fixed points is determined by analyzing the derivative $\frac{du(t+1)}{du(t)}$ evaluated at these points. A fixed point is stable if \mbox{$\left|\frac{du(t+1)}{du(t)}\right|<1$}, and unstable if \mbox{$\left|\frac{du(t+1)}{du(t)}\right|>1$}. The case \mbox{$\left|\frac{du(t+1)}{du(t)}\right|=1$} is inconclusive and requires further analysis.

\newpage
\section{Proofs of Theorems}
\label{appendix:proofs}

\begin{theorem}
    \label{thm:appendix_theo1}
    Under our discrete QIF neuron model using tdBN to normalize pre-synaptic input $I$, such that $I \sim \mathcal{N}(0, u_{th}^2)$, the membrane potential $u$ follows a distribution with mean $\mu_u \approx  a f(u_{th}, u_1, u_2)$ and variance $\sigma_u^2 \approx u_{th}^2 h(u_{th}, u_1, u_2, a)$ where $\mu_u$ and $\sigma_u^2$ are directly proportional to the functions $f$ and $h$, respectively. The functions $f$ and $h$ are defined as $f(u_{th}, u_1, u_2) = u_{th}^2 + u_1u_2$ and $h(u_{th}, u_1, u_2, a) = 1 + a^2 (2u_{th}^2 + (u_1 + u_2)^2)$.
\end{theorem}

\textit{Proof. }
We define the discrete QIF neuron model as
\begin{align}
    u(t+1) &= a(u(t)-u_1)(u(t)-u_2) + I(t),
\end{align}

where $t$ is the timestep, $u$ is the membrane potential, $a$ is a sharpness parameter, $u_1$ and $u_2$ are calculated using \eqref{eq:appendix_QIF_param_calc}, and $I$ is pre-synaptic input. Considering the membrane potential $u(t)$ and assuming the last firing time was $t' < t$, we have
\begin{align}
    \label{eq:proof1_eq_2}
    u(t+1) \approx \sum_{k=t'+1}^{t} a^{t-k-1}(I(k-1)-u_1)(I(k-1)-u_2) + I(k).
\end{align}
This approximation only holds if $a$ is a relatively small constant. In our work, $a$ is typically set to $0.25$. Small values of $a$ ensure that input into the neuron more than two timesteps ago has a minuscule impact on the voltage at timestep $t+1$, meaning we can approximate~\eqref{eq:proof1_eq_2} as
\begin{align}
    \label{eq:proof1_eq_3}
    u(t+1) \approx a(I(t-1)-u_1)(I(t-1)-u_2) + I(t).
\end{align}
Then, under the tdBN assumption that $I \sim \mathcal{N}(0, u_{th}^2)$ and assuming that $I(t)$ is an independent and identically distribution sample (i.i.d) for all $t$, we can approximate the expectation of $u(t+1)$ as
\begin{align*}
    \E[u(t+1)] &\approx \E[a(I(t-1)-u_1)(I(t-1)-u_2) + I(t)] \\
               &= \E[a(I(t-1)^2-I(t-1)(u_1 + u_2) + u_1u_2) + I(t)] \\
               &= \E[aI(t-1)^2] - \E[aI(t-1)(u_1 + u_2)] + \E[au_1u_2] + \E[I(t)]\\
               &= a\E[I(t-1)^2] - a(u_1 + u_2)\E[I(t-1)] + au_1u_2 \\
               &= au_{th}^2 + au_1u_2 \\
               &= a(u_{th}^2 + u_1u_2).
\end{align*}
Likewise, we can approximate the variance of $u(t+1)$ as
\begin{align*}
    \mathrm{Var}[u(t+1)] &\approx \mathrm{Var}[a(I(t-1)-u_1)(I(t-1)-u_2) + I(t)] \\
                &= \mathrm{Var}[a(I(t-1)^2-I(t-1)(u_1 + u_2) + u_1u_2) + I(t)] \\
                &= \mathrm{Var}[aI(t-1)^2] + \mathrm{Var}[aI(t-1)(u_1 + u_2)] + \mathrm{Var}[au_1u_2] + \mathrm{Var}[I(t)] \quad \text{(i.i.d)} \\
                &= a^2 \mathrm{Var}[I(t-1)^2] + a^2(u_1 + u_2)^2 \mathrm{Var}[I(t-1)] + \mathrm{Var}[I(t)] \\
                &= 2 a^2 u_{th}^4 + a^2(u_1 + u_2)^2 u_{th}^2 + u_{th}^2 \\
                &= u_{th}^2(1 + a^2(2u_{th}^2 + (u_1 + u_2)^2)).
\end{align*}
Therefore, we can define functions $f: \R^3 \to \R$ and $h: \R^4 \to \R$ as
\begin{align*}
    f(u_{th}, u_1, u_2) &= u_{th}^2 + u_1u_2 \\
    h(u_{th}, u_1, u_2, a) &= 1 + a^2(2u_{th}^2 + (u_1 + u_2)^2).
\end{align*}
Then $u$ is a distribution with mean $\mu_u \approx af(u_{th}, u_1, u_2)$ and variance $\sigma_u^2 \approx u_{th}^2 h(u_{th}, u_1, u_2, a)$.
\qed


\end{document}